\ifdefined\pdfminorversion\pdfminorversion=7\fi
\documentclass{robocop_preprint}
\usepackage{amssymb,amsmath}
\usepackage{wrapfig}
\usepackage{capt-of}
\usepackage{algorithm}
\usepackage{algpseudocode}
\usepackage{enumitem}\usepackage{siunitx}
\titleformat*{\paragraph}{\normalfont\sffamily\bfseries}
\newcommand{\abbv}{{\rmfamily\textsc{Robo-COP}}}
\title{Co-Evolving Robot Orchestrators and Policies through Deployment}
\author[1]{Xilun Zhang}
\author[1]{Maggie Wang}
\author[1]{Erik Bauer}
\author[1]{Hong-Xing Yu}
\author[1,\dagger]{Huang Huang}
\author[1,\dagger]{Jiajun Wu}
\author[1,2,\dagger]{Marco Pavone}
\renewcommand{\authorlist}{%
  \authorformat[1]{Xilun Zhang}, \authorformat[1]{Maggie Wang},
  \authorformat[1]{Erik Bauer}, \authorformat[1]{Hong-Xing Yu}\\[3pt]
  \authorformat[1,\dagger]{Huang Huang}, \authorformat[1,\dagger]{Jiajun Wu},
  \authorformat[1,2,\dagger]{Marco Pavone}%
}
\affiliation[1]{Stanford University}
\affiliation[2]{NVIDIA Research}
\contribution[\dagger]{Equal advising}
\hypersetup{
 pdftitle={Co-Evolving Robot Orchestrators and Policies through Deployment},
 pdfauthor={Xilun Zhang, Maggie Wang, Erik Bauer, Hong-Xing Yu, Huang Huang, Jiajun Wu, Marco Pavone},
 pdfsubject={Research preprint}
}
\abstract{
Vision-language-action (VLA) policies trained on large datasets are capable within their training domains, yet they still fail to generalize to the variety of situations a robot meets in real-world deployment.
Agentic robot systems complement the policy with a vision-language model (VLM) orchestrator that learns when to call the policy, how to instruct it, and when to use scripted skills instead.
However, because the harness is built around a frozen policy that has limited language steerability, the orchestrator can avoid the policy's failures but never overcome them.
The policy becomes the bottleneck of the whole system.
Fine-tuning the policy can remove this bottleneck, but updating it alone decouples it from an orchestrator tuned to its old behavior.
We propose \abbv{}, in which the orchestrator and policy co-evolve during deployment.
\abbv{} curates skill demonstrations from its own executions, fine-tunes the policy when this data can address recurring failures, and adopts each new policy only after it improves the skills it was trained for.
Across ten simulated RoboLab tasks, \abbv{} raises mean held-out success from 64.8\% to 73.8\% over the same harness with a frozen policy, while fine-tuning on a fixed schedule without verification reaches only 65.8\%. On three real-world tasks, \abbv{} raises held-out success from 38.3\% to 50.0\%.
\abbv{} turns deployment into a self-improving flywheel in which robots learn by doing, with each improvement in execution producing better data for the next round of learning.
Videos and code are available at \url{https://robo-cop.pages.dev/}.
}

\begin{document}
\maketitle

\section{Introduction}
\label{sec:intro}
Robot foundation models trained on large datasets now generalize to many new objects and scenes \citep{openvla, pi0, pi05, droid}.
Still, they degrade under distribution shift once deployed \citep{liberoplus, radar}, and no dataset can cover every situation a robot will encounter.
Language models offer another route: they generalize zero-shot and can write code that plans over motion primitives \citep{saycan,cap,robotool,zeroshotplanner}, but they are limited by the fixed set of primitives they can call.
Recent agentic systems combine the two, with a language model reasoning and orchestrating vision-language-action (VLA) policies for low-level control \citep{pigey,roboharness,harnessvla}.

Such systems also improve by reflecting on failures during deployment \citep{reflect,droc,aspire} and learning when to delegate to a VLA policy \citep{pigey,roboharness,harnessvla}.
Over many episodes, the harness adapts to the current policy, learning which parts of a task to give the policy and which to handle with scripted skills.
The orchestrator can rephrase instructions or add supporting motions, but VLAs have limited language steerability \citep{liberoplus,cover}, so the harness can only route around the policy's failures rather than fix them, and the upper bound of performance remains dictated by the policy.

Updating the policy from deployment experience could raise this ceiling \citep{eximo}, but doing so during deployment is hard.
We study a deployment setting in which the robot must learn while attempting its assigned tasks, without a separate exploration phase.
\textit{The training data must therefore come from these task attempts.}
In a novel environment many of these attempts fail, yet a failed episode often contains correct steps, such as a block that was grasped correctly before it was dropped, so keeping only successful episodes wastes much of the little data available.
Even with new training data, \textit{fine-tuning does not guarantee improvement}: the new checkpoint may perform worse than the deployed policy. Testing it requires further task attempts, so a poor update can cost deployment successes.
Finally, \textit{updating the policy alone leaves the orchestrator behind}: its memory, such as advice to grasp a block with scripted motion because earlier VLA grasps failed, was formed around the old checkpoint and no longer matches the new policy.
Prior harnesses assume a fixed policy \citep{pigey,roboharness} and never face these problems.
We therefore ask: \textbf{how can the orchestrator and its policy co-evolve during deployment?}

\begin{wrapfigure}{r}{0.48\textwidth}
\vspace{-\intextsep}
\centering
\includegraphics[width=\linewidth]{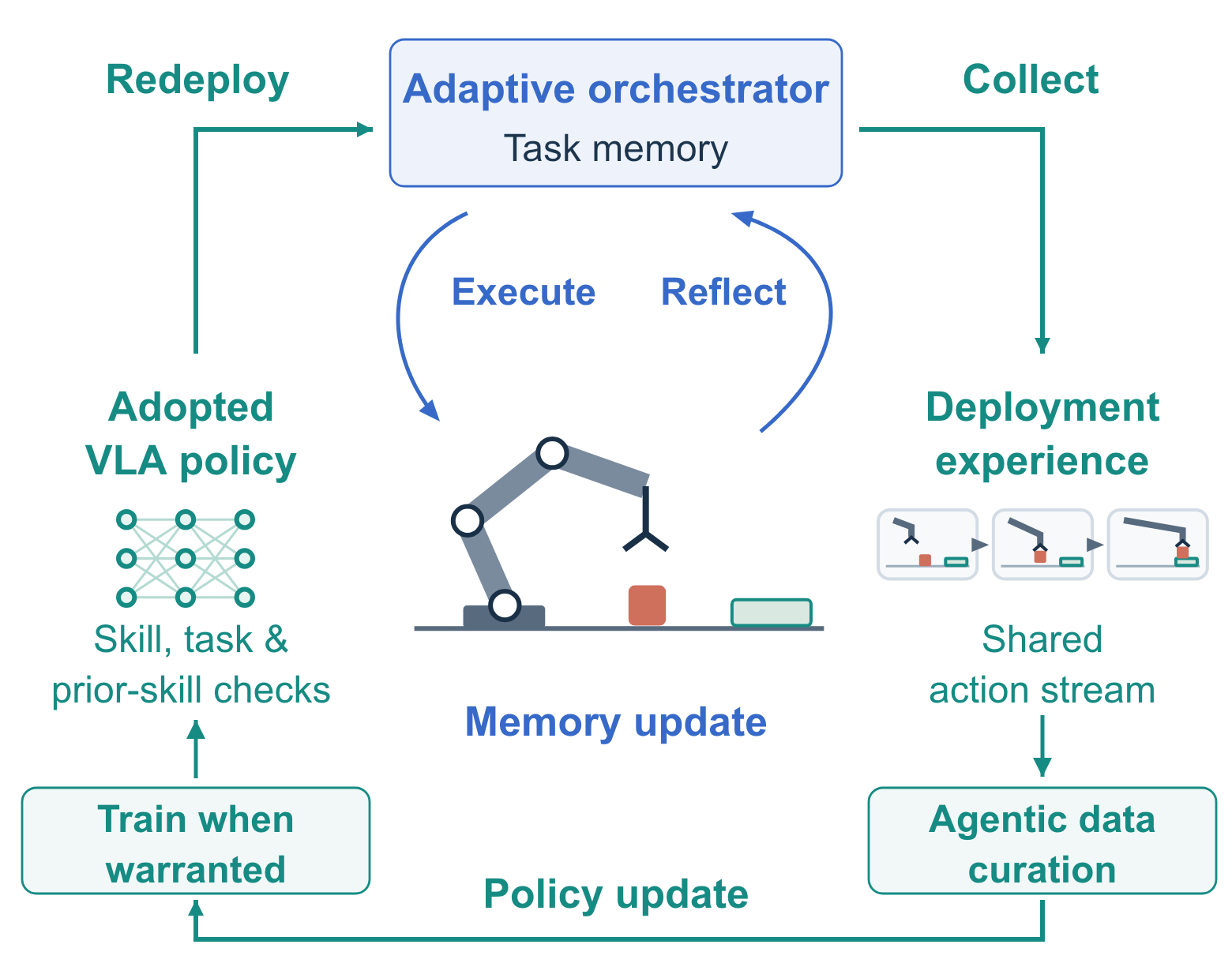}
\caption{\textbf{Two coupled loops for deployment-time improvement.}
The inner loop adapts the harness through episode reflection.
The outer loop uses collected experience to update the policy, then revises memory so the harness can reassess the policy's capabilities.}
\label{fig:nested_loops}
\vspace{-2\intextsep}
\end{wrapfigure}

We introduce \abbv{} (Co-evolving Orchestrator and Policy), a self-improving system whose orchestrator uses scripted motion primitives and a VLA policy as tools, and which updates the policy and adapts the orchestrator's memory together purely during deployment (Figure~\ref{fig:nested_loops}).
It turns its own deployment executions into new training data for the policy by splitting each episode into skills and keeping every skill that succeeded, even when the episode as a whole failed, and it decides when this data warrants training.
After fine-tuning, it evaluates each candidate during deployment, checking its target skills, overall task success, and previously admitted skills before deciding whether to adopt it or retain the current policy. We call this adoption check \emph{policy verification}.
After policy selection, it revisits memory entries that discouraged using the VLA, such as recommendations to use scripted motion after repeated grasp failures, so the orchestrator can reassess whether that advice still applies.
Unlike skill discovery methods that spend a separate budget on exploration \citep{aspire}, \abbv{} learns entirely during deployment, so every trial is an attempt at the task.
We evaluate on ten simulated tasks chosen to span three regimes: tasks the pretrained VLA solves alone, tasks that scripted primitives solve, and tasks that need both.
We also test the full loop on three real-world tasks.

Our contributions are fourfold:
\begin{itemize}[leftmargin=*,itemsep=1pt,topsep=2pt]
    \item \textbf{Co-evolving orchestrator and policy.} We frame deployment-time learning as improving an agentic harness and its policy together.
   \item \textbf{Agentic data curation.} We turn deployment executions into skill demonstrations without manual labels, keeping successful skills even from failed episodes, and train only when the data warrants it.
    \item \textbf{Skill verification and memory revision.} We verify each policy candidate before adoption and revise policy-dependent memory after every update.
    \item \textbf{Performance gains in simulation and the real world.} \abbv{} is competitive on all three regimes, raising mean held-out success from 64.8\% to 73.8\% in RoboLab simulation and from 38.3\% to 50.0\% on the real-world DROID setup over a frozen-policy harness.
\end{itemize}

\section{Related Work}
\label{sec:related}

\paragraph{Agentic robot orchestration.}
Language-model planners compose motion primitives and learned policies into task-solving programs \citep{zeroshotplanner,saycan,cap,voxposer,robotool}.
More recently developed frameworks also improve during deployment: they revise failed programs and retain reusable skills \citep{aspire,resync,graphaspolicy}, or learn when to delegate to a policy rather than an analytic tool \citep{pigey,volo,roboharness,hivla,retriever}.
All of these adapt how a fixed set of capabilities is used, so the system can do no better on a skill than its policy allows.
\abbv{} keeps this adaptive harness but also improves the policy it calls, and revises the harness's memory whenever that policy changes.

\paragraph{Generating robot-policy data.}
Improving a policy requires data that covers the situations it fails on.
Simulation provides scalable supervision \citep{scenesmith}, but transferring learned behavior to physical robots remains difficult \citep{whatwentwrong,dynamicsprompts,simrlvla}.
Real-world datasets and collection interfaces avoid this gap at the cost of human effort \citep{droid,umi,dexumi}, and demonstration-generation methods stretch a small set of human examples further \citep{mimicgen,momagen}.
Closer to our setting, agents and policies can collect and filter their own experience \citep{demoscore,soar,veritas,enpire}.
\abbv{} draws its data from deployment itself, where executions mix policy and scripted actions, and curates skill-level demonstrations from them, including correct skills performed within failed trials.

\paragraph{Policy improvement during deployment.}
Planner distillation and autonomous collection turn robot executions into policy supervision \citep{scalingup,optimus,robocat,soar}, real-world reinforcement learning updates policies from interaction and reward \citep{mark2024parl,wagenmaker2025dsrl,lei2025rl100}, and VLM-guided exploration and skill acquisition generate training data for adapting VLA policies \citep{eximo,wang2026insightselfguidedskillacquisition}.
These methods typically deploy each updated policy, while test-time verification selects instructions or actions for a policy that stays fixed \citep{cover,veritas}.
\abbv{} instead treats each update as a candidate: it decides when training is warranted, verifies the whole candidate policy before adoption, and revises the memory that the harness built around the previous policy.

\begin{figure}[t]
\centering
\includegraphics[width=\linewidth]{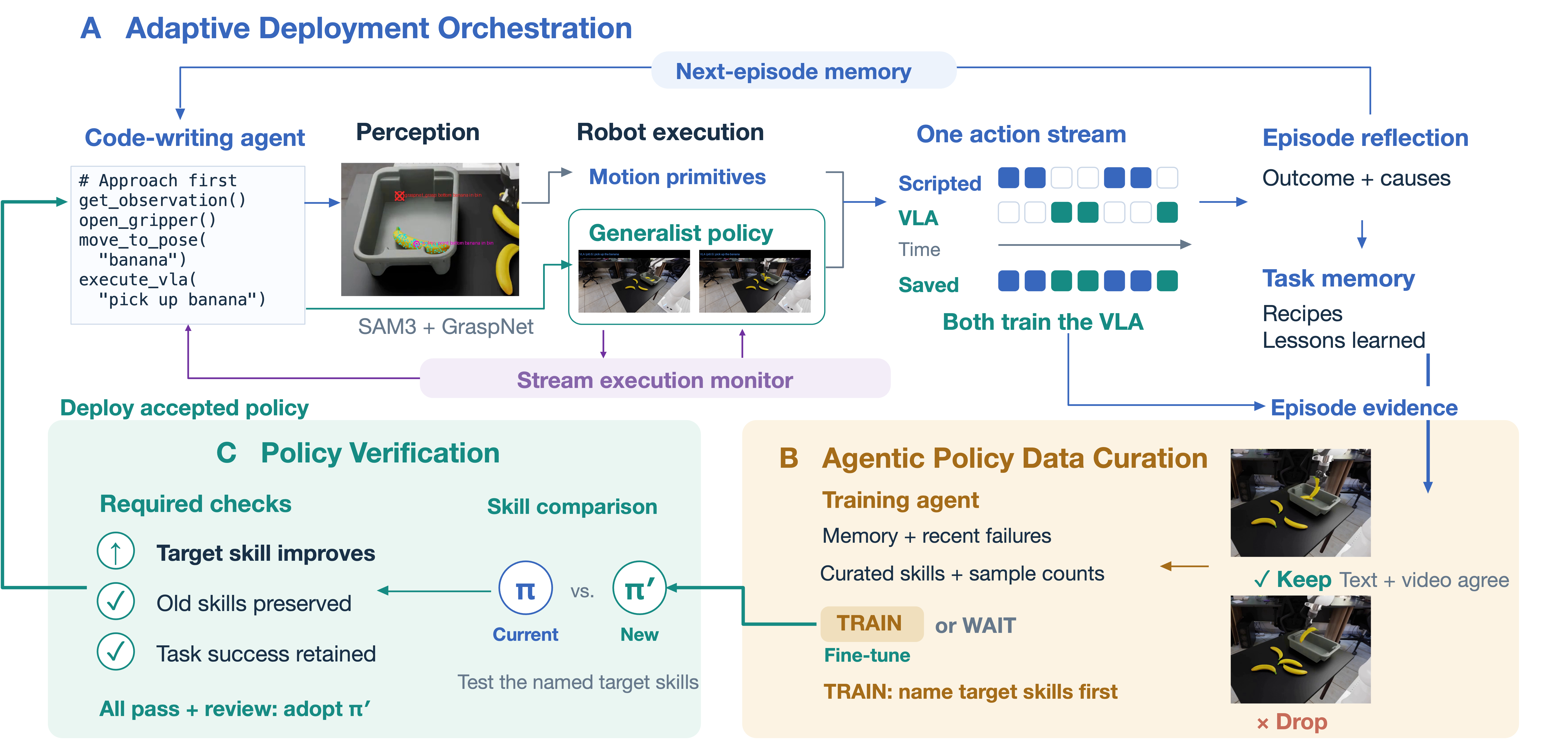}
\caption{\textbf{The \abbv{} deployment-to-learning loop.}
\textbf{A:} The orchestrator mixes scripted motion and VLA calls, logs one action stream, and updates memory.
\textbf{B:} Judges segment the stream into labeled skills; a controller decides when to train.
\textbf{C:} Candidates are verified before adoption, and policy-dependent memory is revised.}
\label{fig:system}
\end{figure}
\section{Method}
\label{sec:method}

\abbv{} updates both the VLA policy and the adaptive harness that orchestrates its use (Figure~\ref{fig:system}).
The harness logs policy and scripted actions as one stream (Section~\ref{sec:orch}), curates it into skill demonstrations (Section~\ref{sec:learn}), and verifies each new policy before adoption, revising the memory tied to the old one (Section~\ref{sec:update}).

\subsection{Adaptive policy orchestration}
\label{sec:orch}

\paragraph{Acting and reflecting.}
\abbv{} is an embodied coding agent built on the CaP-X execution tools \citep{capx}, extended with an \texttt{execute\_vla} tool that invokes the VLA policy on a language instruction.
Its orchestrator writes a program that calls scripted motion primitives, perception functions, and the VLA policy, conditioned on the task goal, current observation, episode history, and task memory $M$, as in prior harnesses \citep{pigey,volo,roboharness}.
A vision-language model (VLM) monitors execution and can interrupt a call or end the episode. Whenever execution stops, a visual judge checks the outcome and either ends the episode or asks for revised code.
It credits the VLA and scripted actions separately, so a successful skill keeps its credit even if a later step fails.
At the end of the episode, the orchestrator records what worked and what failed in $M$, a form of episodic memory that lets agentic robot systems improve without changing any weights \citep{reflect,droc,aspire,roboharness}.

\paragraph{Action unification.}
Training data for the VLA has to come from executions itself, yet many useful motions are produced by scripted primitives rather than by the policy.
Within one execution, the orchestrator may alternate between VLA calls and scripted primitives. 
For example, the robot may approach a red block with scripted primitives and then call the VLA to lower and close the gripper.
We record the commanded joint targets, gripper commands, observations, and robot states at each step in execution order, which yields a continuous action stream in the VLA's training format.
The recorded steps can then become a demonstration labeled \emph{grasp the red block}, even though part of it was produced by scripted primitives.

\subsection{Experience curation and policy improvement}
\label{sec:learn}

\paragraph{From execution to labeled skills.}
The unified action stream contains interleaved VLA and scripted actions, which \abbv{} curates into skill demonstrations $(X,\ell)$, where $X=(o_t,s_t,a_t)_{t=1}^{T}$ is a contiguous trajectory and $\ell$ is its language instruction, and collects the accepted ones in a training set $D$.

\begin{figure}[t]
\centering
\includegraphics[width=\linewidth]{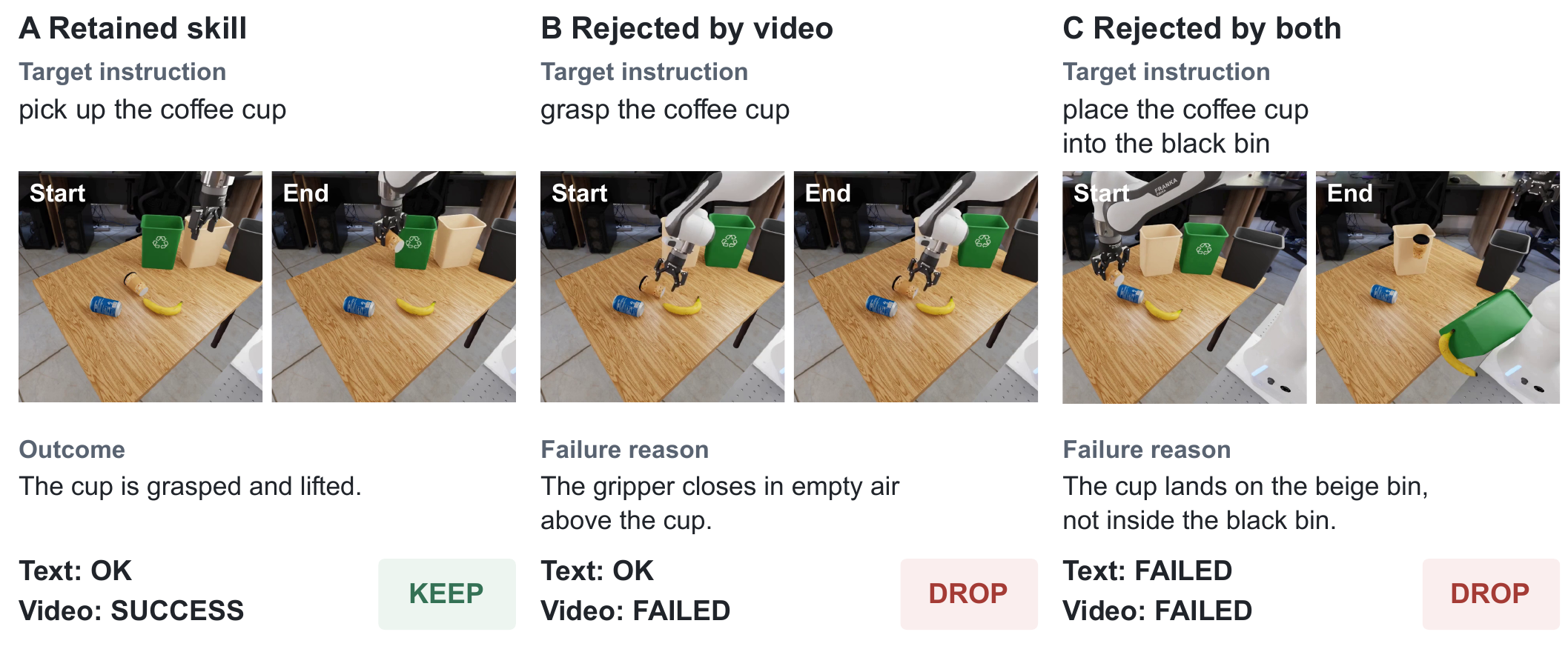}
\caption{\textbf{Skill-level data curation.} Three recorded examples from Recycle cup clutter: both judges accept the pickup (A), the text judge accepts a grasp that the video judge rejects (B), and both reject a placement (C). \textit{The retained pickup comes from an episode that ultimately failed.}}
\label{fig:curation_examples}
\end{figure}

Experience curation evaluates segments independently of the episode outcome, a successful pickup can remain a training example even if the block is later dropped, and a successful episode does not make every segment valid.
For each execution segment, a text judge uses the executed code, console output, and episode feedback to identify each segment's intended skill and assess its outcome.
For VLA calls, it keeps the instruction passed to the policy. 
For scripted primitives, it groups consecutive calls into a motion-level skill, such as the scripted approach above, and a segment labeled \emph{pick up the red block} must also include the lift.
A video judge then checks sampled frames against the proposed instruction summarized from textual judge, and a segment enters $D$ only when the text judge accepts it and the video judge confirms success or clear partial progress.
Figure~\ref{fig:curation_examples} shows a retained pickup and two rejected segments, including a grasp accepted by the text judge but rejected by the video judge.
We therefore expect fine-tuning to improve most on skills that recur across episodes, such as grasping, which the failure analysis in Section~\ref{sec:results} examines.
Appendix~\ref{app:curation} gives the segmentation rules and judge outputs.

\paragraph{Deciding when to train.}
Training on a fixed schedule causes two potential problems: it trains on tasks the harness already solves, where an update can only add risk, and it trains before enough demonstrations have accumulated to support a useful update.
\abbv{} therefore lets a language-model controller decide when to train given the deployment data so far.
It reviews the curated demonstrations alongside recent performance, task memory, failures, and earlier attempts, and returns \textsc{train} or \textsc{wait} with the desired improvement or the evidence still needed.
We expect the controller to skip training on tasks the harness already solves, and to wait on the others until the curated demonstrations cover the skills that keep failing (Section~\ref{sec:results}).

A \textsc{train} decision names up to four target skill instructions $E$, which guide verification. 
Training itself uses all curated data collected so far.
We fine-tune a candidate $\pi'$ from the pretrained checkpoint with LoRA \citep{lora} and the policy's original flow-matching objective.
Deployment on the task pauses while $\pi'$ trains. 
Afterward, $\pi$ stays in service as the baseline against which $\pi'$ is evaluated.
Appendices~\ref{app:controller} and~\ref{app:train} give the controller and training details.

\subsection{Policy verification and integration}
\label{sec:update}

\paragraph{Candidate policy verification.}
We evaluate the candidate $\pi'$ during a window of $W$ trials while retaining the deployed policy $\pi$ as a fallback.
For target skills, the orchestrator first tries $\pi'$ and falls back to $\pi$ or scripted motion if needed.
Because fallbacks can complete a task after $\pi'$ fails, task success alone does not reflect the candidate's own performance, so verification is performed at the skill level rather than only at the episode level.
Since each candidate is fine-tuned from base policy on all curated data, it could also lose a skill that an earlier update added, so a verifier checks for regressions on skills adopted with earlier candidates, which we call \emph{admitted} skills, $S$.

Let $b$ be the incumbent's recent task success rate and $\mathrm{succ}(W)$ the task success rate during the candidate window.
For each admitted skill $s$, let $r_{\mathrm{inc}}(s)$ and $r_c(s)$ be the observed per-call finish rates of the incumbent and candidate, where a call finishes when its completion check fires.
We apply the skill check to $S_{\mathrm{obs}}$, the admitted skills with sufficient candidate calls and an incumbent reference, and require
\begin{equation}
\mathrm{succ}(W) \ge b-\delta_{\mathrm{task}},\qquad
r_c(s) \ge r_{\mathrm{inc}}(s)-\delta_{\mathrm{skill}}
\quad \forall s\in S_{\mathrm{obs}}.
\label{eq:admit}
\end{equation}
If the candidate receives too few calls to assess, verification returns \textsc{starved}, and the candidate is discarded and memory is left unchanged.
If either inequality fails, it returns \textsc{revert} and $\pi$ stays deployed without updating policy weights.

\begin{algorithm}[t]
\caption{\abbv{} Self-Improving Pipeline}
\label{alg:main}
\small
\begin{algorithmic}[1]
\State $\pi\gets\pi_{\mathrm{base}}$; $M,D,S\gets\emptyset$
\While{deployment trials remain}
  \State $\tau\gets\textsc{Execute}(\pi,M)$; \ $M\gets\textsc{Reflect}(M,\tau)$; \ $D\gets D\cup\textsc{Curate}(\tau)$
  \If{$\textsc{Controller}(D,M)$ returns \textsc{train} with targets $E$}
    \State $\pi'\gets\textsc{FineTune}(\pi_{\mathrm{base}},D)$
    \State $v\gets\textsc{Verify}(\pi',\pi,E,S)$ \Comment{runs over the next $W$ trials}
    \If{$v=\textsc{adopt}$} \State $\pi\gets\pi'$; \ $S\gets S\cup E$ \EndIf
    \If{$v\neq\textsc{starved}$} \State $M\gets\textsc{ReviseMemory}(M)$ \EndIf
  \EndIf
\EndWhile
\end{algorithmic}
\end{algorithm}

For the target skills $E$, a language-model reviewer checks what each candidate call actually achieved, given the generated code, output logs, and visual judge outputs.
Verification does not rely on episode success alone but checks whether the skills actually meet the expectations.
A candidate is adopted only if it satisfies all requirements, and otherwise the system reverts to the current policy.
Adoption is decided only at the end of the evaluation window, whereas reversion can happen earlier if the candidate clearly fails during evaluation.
Appendix~\ref{app:adopt} specifies the window size, thresholds, and stopping rules.

\paragraph{Policy adoption and memory revision.}
A candidate that passes verification replaces $\pi$, and its target skills are added to $S$. 
If no candidate is ever adopted, the system keeps the base policy and improves through orchestration alone.
Changing the deployed policy does not change the advice stored in task memory.
For example, memory may recommend scripted motion for grasping a red block because earlier VLA attempts failed. Following that recommendation after an update would keep the harness from discovering that the new policy can successfully grasp.
The same problem arises when a candidate is rejected.

During its evaluation window, episode reflection records the candidate's failures as advice against using the VLA. After the incumbent is restored, that advice would wrongly steer every later episode away from it.
After adoption or reversion, we therefore rewrite references to the candidate to name the policy in service and mark advice that steers the orchestrator away from the VLA as unverified for the current checkpoint.
The advice is marked rather than deleted, since the update may not have fixed the skill, so later episodes can retest it while its history stays in memory.
Algorithm~\ref{alg:main} summarizes the full loop.

\section{Experiments}
\label{sec:exp}
Through this section, we aim to address three main questions.
\textbf{Q1:} Does co-evolution improve held-out success over a frozen-policy harness and fixed-schedule training, and in which task regimes?
\textbf{Q2:} Does the system decide correctly when to train and whether to adopt?
\textbf{Q3:} Does learning during deployment cost task success along the way?
We answer them on ten simulated tasks and then test the full loop on three real-world tasks.

\subsection{Experimental setup}
\label{sec:setup}
\label{sec:protocol}

\begin{figure}[!htbp]
\centering
\includegraphics[width=\textwidth]{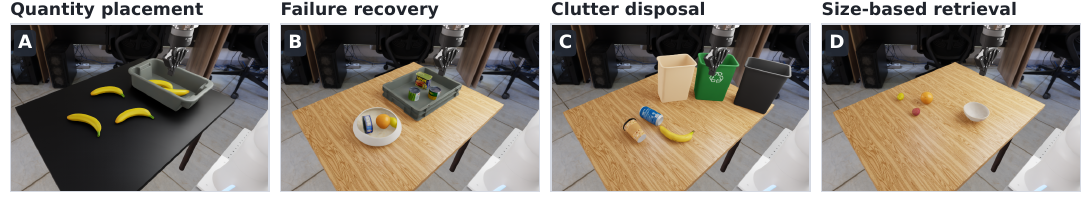}
\caption{\textbf{Representative RoboLab tasks.} Object configurations vary across deployment and held-out evaluation.}
\label{fig:simulation_tasks}
\end{figure}

\paragraph{Tasks and system.}
We evaluate on ten RoboLab~\citep{RoboLab} tasks with randomized object configurations, spanning retrieval, grouping, placement, and recovery.
We selected them before running \abbv{} to represent the three regimes introduced in Section~\ref{sec:intro}.
Within every regime, the controller ends up waiting on one task, adopting a candidate on another, and reverting at least one candidate (Table~\ref{tab:first_training_turn}), so the results cover each decision it can make.
Figure~\ref{fig:simulation_tasks} shows representative scenes, and Appendix~\ref{app:task_visuals} shows successful episodes and randomized resets for all ten tasks.
The initial VLA is $\pi_{0.5}$ trained on DROID~\citep{pi05,droid} and served at its training control rate, and Gemini 3.8 Flash runs the orchestrator.
Appendices~\ref{app:api} and~\ref{app:train} give the execution tools and fine-tuning settings.

\paragraph{Baselines.}
\emph{$\pi_{0.5}$ only} runs the base policy without an orchestrator, memory, or updates.
\emph{Frozen-policy baseline} uses the same harness and episode reflection as \abbv{} but never changes the VLA weights, isolating what memory adaptation alone achieves.
\emph{Fixed schedule} uses the same recorder, curator, fine-tuning procedure, episode reflection, and memory revision as \abbv{}, but trains every 30 trials and deploys each checkpoint without verification, isolating the contribution of the controller and verifier.
\emph{ASPIRE} diagnoses failures, revises robot programs, and retains validated solutions as reusable skills~\citep{aspire}.
Since ASPIRE did not evaluate on RoboLab, we adapt its algorithm to our benchmark and API setting.

\paragraph{Protocol and metrics.}
Frozen-policy baseline, fixed schedule, and \abbv{} each run 100 deployment trials (episodes) per task on matched initializations, sharing their first 30 trials before any training decision.
\emph{Deployment success} is the success rate over these trials, which include data collection and candidate evaluation.
We then freeze each system's policy and memory, or ASPIRE's skill library after its 100-execution search, and measure \emph{held-out success} on 50 held-out initializations per task, shared across all methods and distinct from those used during deployment, following the single-learning-run protocol of prior agentic robot-learning work \citep{aspire,enpire}.
The standalone $\pi_{0.5}$ is tested only on the held-out trials.

\subsection{Simulation results}
\label{sec:results}

\begin{figure}[!htbp]
\centering
\includegraphics[width=\linewidth]{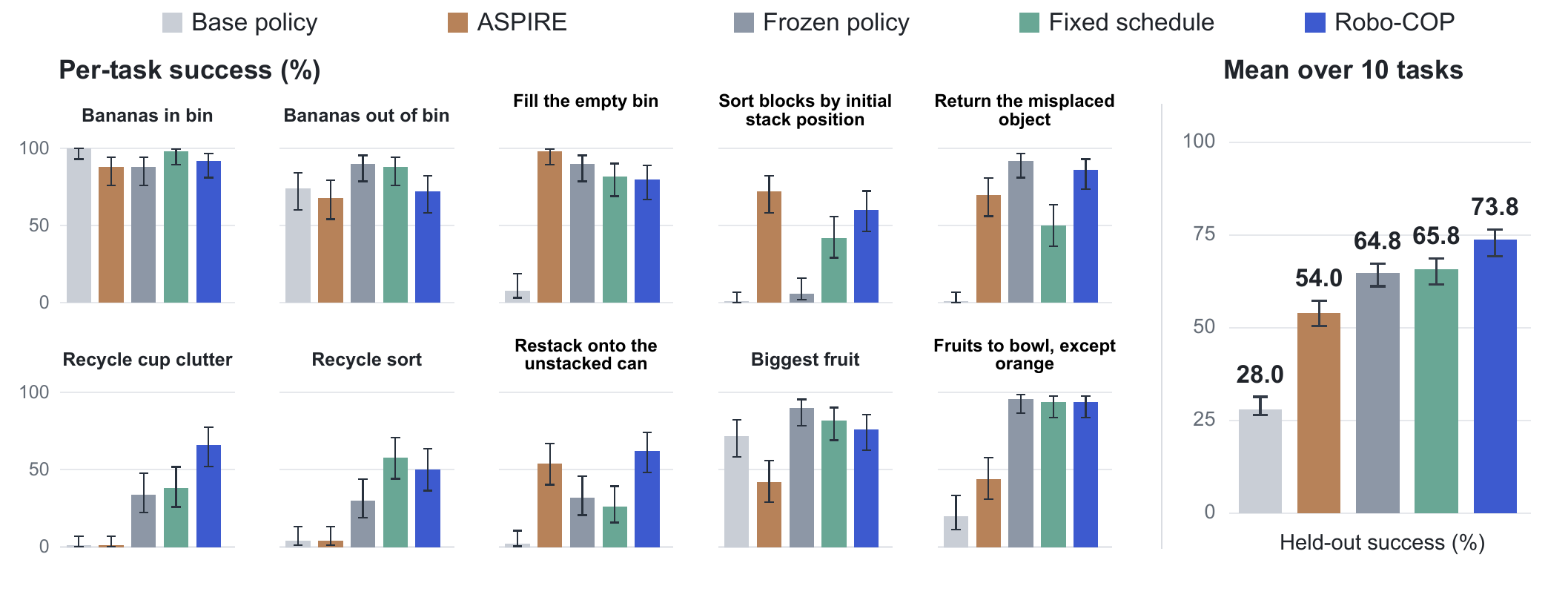}
\caption{\textbf{Held-out success on 50 shared initializations per task.} Left: per-task success for all five methods. Right: equally weighted ten-task means; whiskers are 95\% beta-posterior credible intervals over the fixed task set (Appendix~\ref{app:simulation_results}).}
\label{fig:heldout_main}
\end{figure}

\paragraph{Q1: Does co-evolution improve held-out success, and where?}
Co-evolution improves held-out success over both baselines, and the gains are largest when \abbv{} updates the policy.
\abbv{} reaches 73.8\% mean held-out success, versus 64.8\% for the frozen-policy baseline and 65.8\% for the fixed schedule, which recovers only 1.0 of the 9.0 points (Figure~\ref{fig:heldout_main}).
The standalone VLA solves every held-out trial of Bananas in bin but succeeds on at most 8\% of trials on six tasks.
The gain concentrates where the policy changed: on the six tasks where \abbv{} adopted an update, it averages 11.3 points over the frozen-policy baseline, versus 5.5 points on the four tasks where it kept the base policy, where the systems differ only in their memory and deployment histories.
Omitting any single task leaves a mean gain of 4.0--12.0 points over the frozen-policy baseline and 4.9--10.7 over the fixed schedule (Figure~\ref{fig:task_robustness}).

\begin{table}[!htbp]
\centering
\small
\setlength{\tabcolsep}{4pt}
\caption{\textbf{Held-out success by task regime (\%).}
Tasks are grouped by whether they can be completed using the pretrained VLA alone (Bananas in bin, Bananas out of bin, Biggest fruit), scripted primitives alone (Fill the empty bin, Sort blocks by initial stack position, Return the misplaced object, Restack onto the unstacked can), or neither (Recycle cup clutter, Recycle sort, Fruits to bowl, except orange).
The last three columns count tasks by \abbv{}'s final policy decision.}
\label{tab:regimes}
\begin{tabular}{@{}lcrrrrrccc@{}}
\toprule
Regime & Tasks & $\pi_{0.5}$ & ASPIRE & Frozen & Fixed & Ours & No training & Adopted & Reverted (final) \\
\midrule
VLA alone & 3 & 82.0 & 66.0 & \textbf{89.3} & \textbf{89.3} & 80.0 & 1 & 2 & 0 \\
Primitives alone & 4 & 2.5 & \textbf{73.5} & 55.0 & 50.0 & 72.0 & 1 & 2 & 1 \\
Neither alone & 3 & 8.0 & 16.0 & 53.3 & 63.3 & \textbf{70.0} & 1 & 2 & 0 \\
\bottomrule
\end{tabular}
\end{table}

The regimes separate the methods as anticipated in Section~\ref{sec:intro} (Table~\ref{tab:regimes}). 
Each method leads where its adaptation matches the task, and \abbv{} is the only one that stays competitive in all three regimes.
Where the VLA alone suffices, every harness that can call the policy stays close to $\pi_{0.5}$ (82.0\%): the frozen and fixed harnesses reach 89.3\% and \abbv{} 80.0\%, while ASPIRE, which cannot call the VLA, falls to 66.0\%.
Where scripted primitives suffice, ASPIRE leads (73.5\%) because it searches over programs built from exactly those primitives, and \abbv{} comes close (72.0\%), while the frozen and fixed harnesses reach 55.0\% and 50.0\%.
Where neither suffices, ASPIRE falls to 16.0\% because its programs cannot call the VLA, while \abbv{} fine-tunes the missing skills on the two tasks that need them and reaches 70.0\%, 16.7 points above the frozen-policy baseline.

Across regimes, the skill that fine-tuning improves most is grasping.
\abbv{} fails 131 of 500 held-out trials versus 176 for the frozen-policy baseline, and grasp or drop, the dominant failure in both (40 and 68), accounts for 28 of the 45 fewer failures: every first training request targeted a pickup skill (Table~\ref{tab:first_training_turn}), and fine-tuning on curated grasps, including those from failed episodes, makes the policy's grasps more reliable.

\begin{wrapfigure}{r}{0.46\textwidth}
\centering
\includegraphics[width=\linewidth]{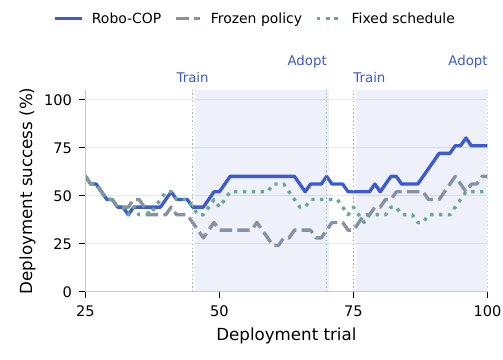}
\caption{\textbf{Deployment success on Recycle cup clutter.} Success rates use a trailing 25-trial window, starting at trial 25, to reduce fluctuations from randomized scenes and stochastic agent decisions and show longer-term deployment trends. Shading marks candidate evaluation; labels mark training requests and adoption decisions.}
\label{fig:cup_decisions}
\end{wrapfigure}

\paragraph{Q2: Does the system decide correctly when to train and whether to adopt?}
A good controller should track the better of the two baselines on each task: train when training helps, as the fixed schedule does, and hold back when it does not, as the frozen-policy baseline does.
The controller trained on the seven tasks where the harness still failed often, with the frozen-policy baseline succeeding on 45--79\% of deployment trials, and waited on the three where it rarely failed (80--96\%), as its prompt directs (Appendix~\ref{app:decision_prompts}).
On those three tasks, the fixed schedule's nine training runs bring no net gain over the frozen-policy baseline.
In every first training request (Table~\ref{tab:first_training_turn}), the targets include a pickup skill, which matches the grasp and drop failures that dominate held-out errors in the failure analysis.
Verification then rejected 5 of 12 candidates.
On the four tasks whose first candidate was reverted, the fixed schedule, which deploys every checkpoint, falls below the frozen-policy baseline on three (Table~\ref{tab:sim_full}), so the rejections avoided real regressions.

Rejection also does not end learning: the controller waits for new demonstrations and trains again, and on three of those four tasks the next candidate was adopted (Table~\ref{tab:sim_full}).

Figure~\ref{fig:cup_decisions} follows two updates on Recycle cup clutter. The controller requests training after trials 45 and 75 and adopts the candidates at trials 70 and 100. Success fluctuates after the first update, then rises during the second candidate-evaluation window, reaching 76\% over the final 25 deployment trials, compared with 60\% for the frozen-policy baseline and 52\% for the fixed schedule.

\textbf{Q3: Does learning cost deployment success?}
\abbv{} learns without sacrificing success during deployment.
It succeeds on 73.2\% of 100 deployment trials, versus 68.7\% for the frozen-policy baseline and 70.3\% for the fixed schedule.
The gap appears once the runs diverge after trial 30 (72.9\% versus 66.4\% and 68.7\% over trials 31--100) and holds over the final 25 trials (73.6\% versus 67.6\% and 67.2\%) and every other late window we examined (Appendix~\ref{app:simulation_results}).
ASPIRE spends its 100-execution budget exploring rather than attempting the task, so it has no comparable deployment success.

\subsection{Real-world experiments}
\label{sec:real}

\begin{figure}[!htbp]
\centering
\begin{minipage}[t]{0.32\linewidth}\vspace{0pt}
\centering\includegraphics[width=\linewidth]{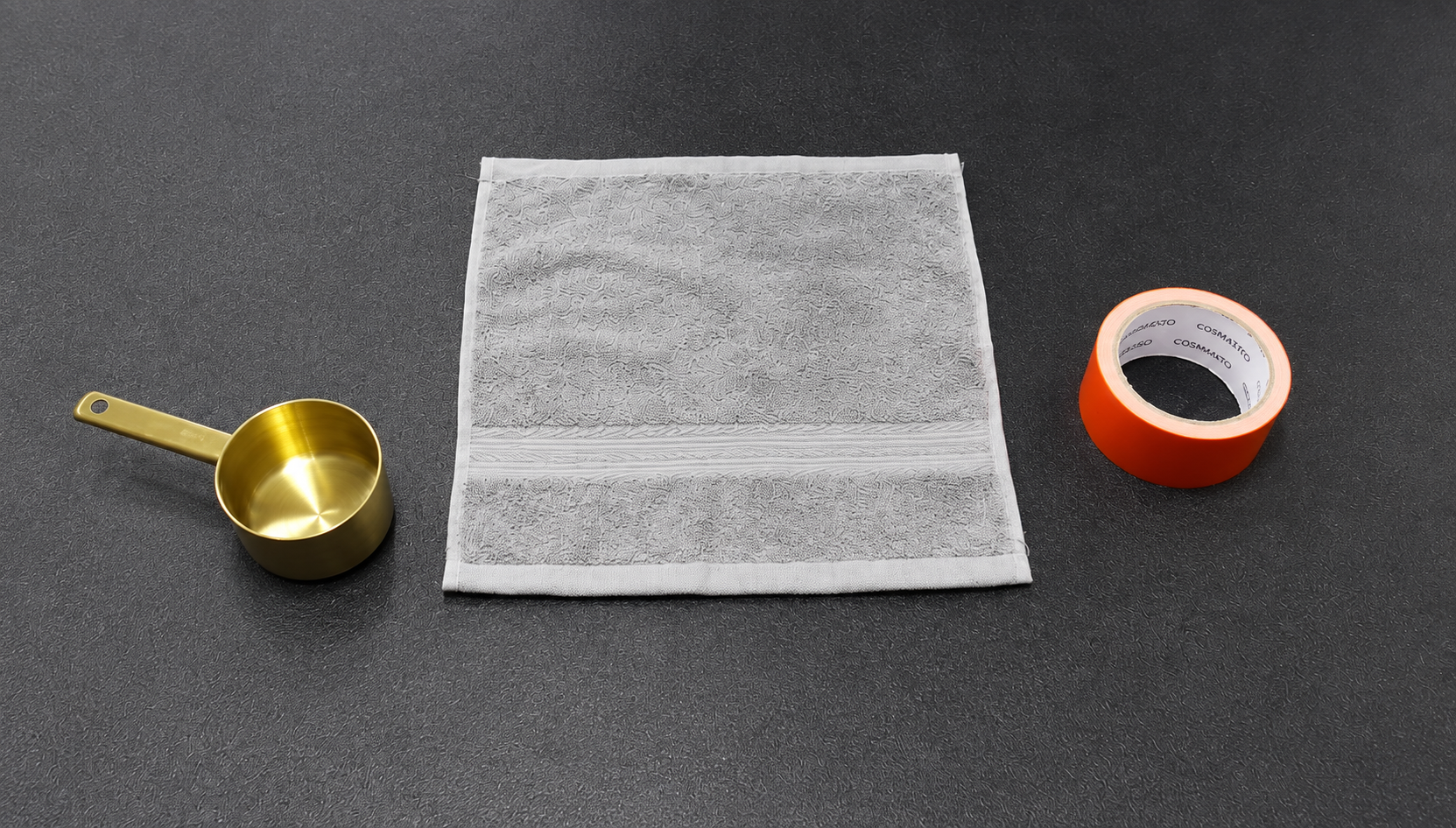}\par\small\emph{Put the cup and tape on the towel} (\SI{40}{\second})
\end{minipage}\hfill
\begin{minipage}[t]{0.32\linewidth}\vspace{0pt}
\centering\includegraphics[width=\linewidth,trim=40bp 0 40bp 45bp,clip]{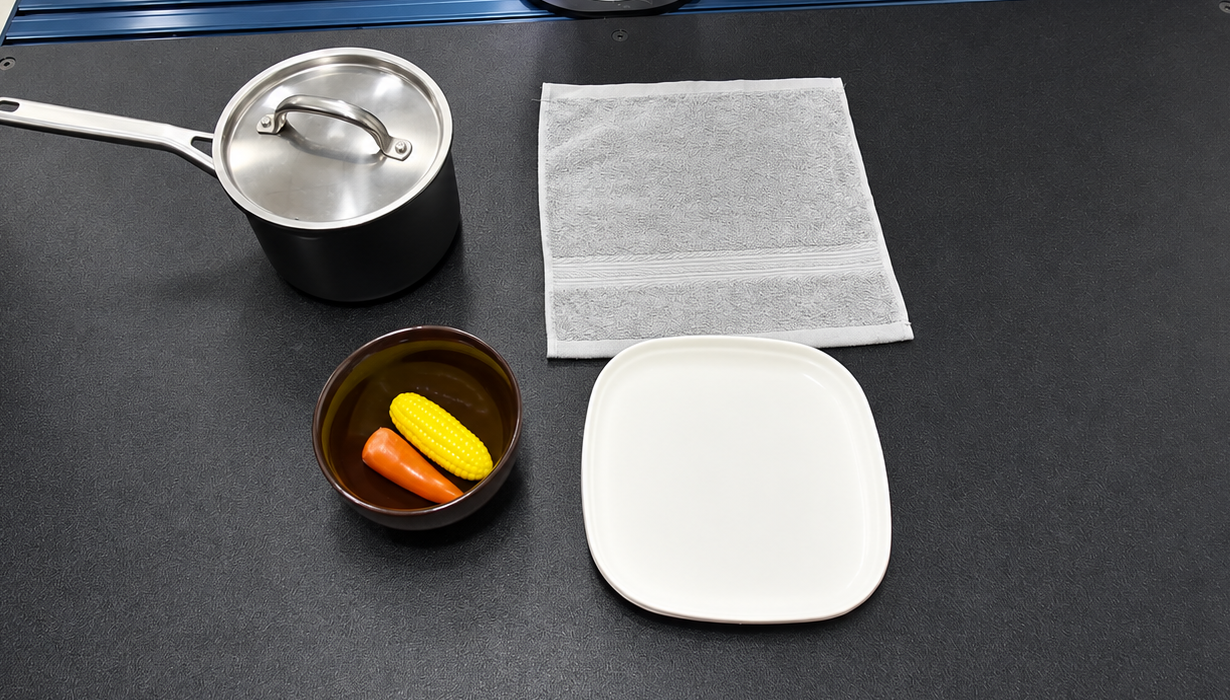}\par\small\emph{Move the lid to the towel and the contents to the plate} (\SI{60}{\second})
\end{minipage}\hfill
\begin{minipage}[t]{0.32\linewidth}\vspace{0pt}
\centering\includegraphics[width=\linewidth]{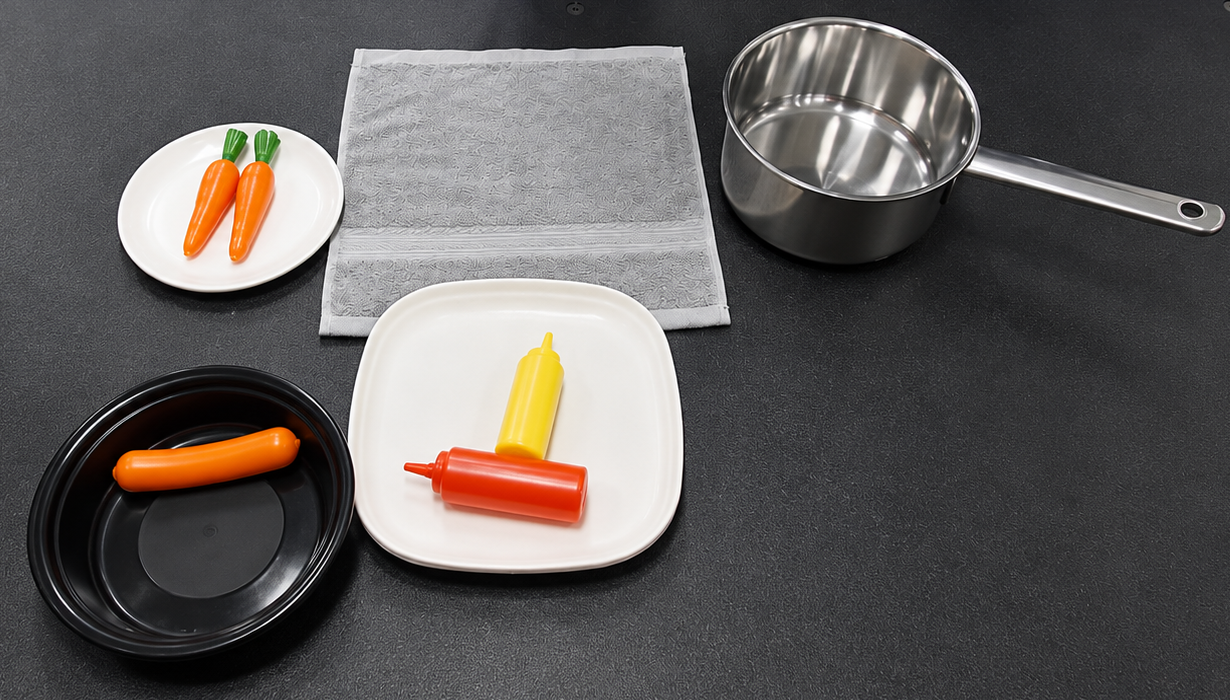}\par\small\emph{Put the mustard in the bowl and the carrots in the pot} (\SI{90}{\second})
\end{minipage}
\caption{\textbf{Real-world tasks on a DROID setup.} Each task requires placing multiple objects, and a trial succeeds only if every condition is met within the time limit shown.}
\label{fig:real_world_tasks}
\end{figure}

We run the same protocol on three physical tasks with 40-, 60-, and 90-second limits (Figure~\ref{fig:real_world_tasks}), with 50 learning trials and 20 held-out tests per task, comparing \abbv{} with a frozen-policy baseline that uses the same policy, tools, and memory adaptation.
Because each task has only 50 learning trials, we shorten the adoption window to $W=10$ trials.
A trial succeeds only if every requested condition is met within its episode time budget. Appendix~\ref{app:real_world} shows successful executions.

\begin{table}[!htbp]
\centering
\caption{\textbf{Real-world success rates (\%) and \abbv{}'s training decisions.} Frozen uses the same harness but never updates the policy.}
\label{tab:real_world_results}
\small
\setlength{\tabcolsep}{4pt}
\begin{tabular}{@{}lcccccc@{}}
\toprule
& \multicolumn{2}{c}{\abbv{} decisions} & \multicolumn{2}{c}{Learning (50 trials)} & \multicolumn{2}{c}{Test (20 trials)} \\
\cmidrule(lr){2-3}\cmidrule(lr){4-5}\cmidrule(l){6-7}
Task & Train request & Outcome & Frozen & \abbv{} & Frozen & \abbv{} \\
\midrule
Cup and tape (\SI{40}{\second})           & Train@30 & Revert & 46 & 44 & 40 & 40 \\
Lidded pot (\SI{60}{\second})             & Train@30 & Adopt  & 32 & 38 & 35 & 70 \\
Mustard and carrots (\SI{90}{\second}) & Wait     & n/a    & 32 & 32 & 40 & 40 \\
\midrule
Mean & & & 36.7 & \textbf{38.0} & 38.3 & \textbf{50.0} \\
\bottomrule
\end{tabular}
\end{table}

\paragraph{Results.}
The real-world results reproduce each controller outcome seen in simulation, and only the verified update changes held-out success (Table~\ref{tab:real_world_results}).
On the 60-second task, the controller trains at trial 30, verification adopts the candidate, and held-out success doubles from 35\% to 70\%.
On the 40-second task, the controller also trains at trial 30, but the candidate fails on several evaluation trials. 
Verification reverts it and deployment continues with the incumbent, so learning (44\% versus 46\%) and held-out success (40\% versus 40\%) stay level with the frozen-policy baseline, and the rejected update costs little.
On the 90-second task, the controller never trains, so both systems share one learning trace and reach 40\% held-out.
A separate diagnostic that forced training on this task and deployed the checkpoint without verification reached only 20\% held-out, half of either system, confirming that waiting was the right decision.
Across the three tasks, \abbv{} raises mean held-out success from 38.3\% to 50.0\%.

\section{Conclusion}
\label{sec:limits}
\abbv{} co-evolves a robot's orchestrator and policy during deployment, turning the deployment experience into steady improvement of the system.
Across ten simulated manipulation tasks, \abbv{} improves held-out success by 9.0 points over the same harness with a frozen policy, while fine-tuning on a fixed schedule without verification recovers only 1.0 of those points.
The controller left the policy unchanged on three tasks where the frozen-policy baseline already achieved 80--96\% deployment success. On the remaining seven tasks, it requested training, but verification rejected 5 of 12 candidates rather than deploying every fine-tuned checkpoint.
In the real world, \abbv{} improves performance on the task where it adopts an update, loses little on the task where it reverts, and forcing an update the controller rejected hurts performance. 

\paragraph{Limitations.}
\abbv{} can learn only skills that its policy and scripted tools already demonstrate at least partially, since every training example comes from its own executions.
Physical deployment requires human resets, and each policy update adds curation, training, and evaluation cost.
Where scripted primitives already cover a task, ASPIRE's program search outperforms \abbv{} (Fill the empty bin and Sort blocks by initial stack position), because \abbv{}'s orchestrator adapts through memory rather than structurally searching over programs.
Finally, even though all design components have been verified through experiments, the scalability of \abbv{} still needs to be validated on larger experiments to confirm that the results hold at scale.

\paragraph{Future work.}
\abbv{} is a step toward robot systems that improve themselves in the real world through deployment.
A fully self-improving system requires every component to improve.
\abbv{} only updates part of the whole system, the policy and the orchestrator's memory, while its scripted skills, its judges, and the orchestrator's own reasoning stay fixed.
Structured skill exploration, as in ASPIRE, diagnoses failures, proposes new programs, and retains the validated ones as reusable skills.
Running such exploration inside the deployment loop, directed at the task at hand, would give the orchestrator better tools whose successful executions become training data for the policy, and the orchestrator itself could be better aligned with reinforcement learning from deployment outcomes.
Because \abbv{} curates demonstrations at the skill level, training a single policy on skills pooled from many tasks is a natural path to cross-task generalization, so that a system deployed long enough improves at every task it encounters.

\bibliographystyle{assets/plainnat}
\bibliography{reference}

\clearpage
\appendix
\section*{Appendix}
\addcontentsline{toc}{section}{Appendix}

\section{Constants}
\label{app:hyper}
All reported simulation results use the settings in Table~\ref{tab:hyper}.

\begin{table}[!hbp]
\caption{Constants defining the loop.}
\label{tab:hyper}
\begin{center}\small
\begin{tabular}{llp{6.2cm}}
\toprule
Constant & Value & Raising it \\
\midrule
Turns per episode            & 20    & more deliberation per episode, fewer episodes per budget \\
Memory budget                & 8000 char & more recall, less of the prompt in the attended region \\
Stillness exit               & 16 steps & fewer premature exits, more budget spent on stalled bursts \\
Held frames per segment      & 15 & stronger stopping behaviour, more static frames in $D$ \\
Action horizon $H$           & 15    & longer commitment per inference, coarser reaction \\
Curated floor / new since last & 30 / 10 & later first training run, more evidence behind it \\
Named instructions per run   & $\le 4$ & broader expectation, weaker per-skill evidence \\
Adoption window $W$          & up to 25 episodes & more evidence, fewer episodes for further updates \\
Candidate-exposed trial floor & 5 & stronger exposure requirement, more starved rounds \\
Prior-skill call floor       & 5 & more evidence per check, fewer prior skills checked \\
Margins $\delta_{\mathrm{task}},\delta_{\mathrm{skill}}$ & 0.1 & more tolerated decline, weaker safeguards \\
Task bar window              & 30 episodes & steadier bar, slower to reflect a real change \\
Attempts per task            & 3     & more chances to improve, more episodes spent on windows \\
\bottomrule
\end{tabular}
\end{center}
\end{table}

\section{Orchestrator interface}
\label{app:api}
The orchestrator's programs run in a sandbox that exposes only the following functions: \texttt{get\_observation}, \texttt{get\_object\_pose}, \texttt{sample\_grasp\_pose}, \texttt{compute\_place\_pose}, \texttt{goto\_pose}, \texttt{goto\_pregrasp}, \texttt{open\_gripper}, \texttt{close\_gripper}, and \texttt{execute\_vla}.
Documentation for each function is generated from its signature and docstring.
Object poses come only from the perception functions; the sandbox has no access to ground-truth poses and blocks imports of simulator packages.
Each episode allows at most twenty turns.

\paragraph{Execution tools and monitoring.}
Scripted execution selects targets with segmentation \citep{sam3}, visual pointing \citep{molmo}, grasp proposals \citep{contactgraspnet}, and placement geometry, and executes them with a collision-aware planner \citep{curobo} and gripper commands.
The policy is called through \texttt{execute\_vla(instruction, max\_steps, policy)}.
It outputs arm and gripper commands.
We refer to a single invocation as a policy call.
A call ends when its completion check fires, when it reaches its step limit, or when the VLM monitor stops it.
An arm that stops moving while still holding an object is not counted as a completed placement.
Program variables persist across turns.

\paragraph{Episode memory and policy reconciliation.}
\label{app:memory}
After each episode, a script logs the outcome, the instruction and exit condition of each policy call, and the scripted grasp poses.
A reflection model then updates two memory sections, \emph{what works} and \emph{known failure modes}, and may propose deleting specific lines from a \emph{turn-verified recipe}.
Deletions are applied by code, so recipe steps are never dropped silently.
Memory is capped at 8000 characters with a fixed allocation per section to keep prompts short \citep{lostinmiddle,ruler}; similar memory designs appear in \citet{aspire,resync}.
After each adoption or reversion, entries that discourage using the policy are marked \textsc{unverified} for the current checkpoint, and references to the candidate are rewritten to name the policy in service.
These entries keep their counts and last-observed episode as history, but the counts no longer set the orchestrator's confidence.
All other entries are left unchanged.

\paragraph{Recording fields.}
The recorder logs every actuated step, whether it comes from the inverse-kinematics servo, the motion planner, a gripper command, or the policy.
Each row stores both camera views, the end-effector pose, the measured joint state, the commanded joint target with an eighth channel for the gripper, and the turn index.

\section{Curation}
\label{app:curation}
Curation runs in two stages: a text judge proposes and labels skills, and a visual judge checks them against video.

\paragraph{Stage A (text).}
For each turn, the text judge reads the code, console output, and outcome assessment, and returns an instruction and a verdict.
Each instruction names one skill using a canonical verb and its required arguments.
When the turn called the policy with an instruction string, that string is kept verbatim.

\paragraph{Stage A (skill grouping).}
For scripted turns, we mark each primitive call in the code and ask the judge to group consecutive primitives into physically meaningful skills, named with the same rule.
Function names may not appear in instructions.

\paragraph{Stage B (video).}
The visual judge samples six frames at native resolution from the segment's video, receives the Stage A instruction, and returns one of five verdicts: success, useful partial progress, no progress, failure, or unclear.

\paragraph{Filtering.}
A segment is kept when Stage A returns \textsc{ok} and Stage B returns success or useful partial progress.
Turns whose only instruction opens or closes the gripper are dropped before Stage B.
Both verdicts are cached with their reasons, so the filter can be retuned offline without new model calls.

\paragraph{Segmentation.}
Accepted rows are split wherever the instruction changes, a rejected row intervenes, or the joint state jumps by more than $0.3$\,rad.
Segments shorter than two steps are discarded.
Each remaining segment is padded with fifteen held frames that repeat the final observation and use the final joint state and gripper command as the action, so that the policy learns to stop at the end of a skill.

\section{Controller}
\label{app:controller}
Before the language model is consulted, code applies several guards.
No decision is made while a candidate is under evaluation.
Training waits until at least thirty episodes contain a kept segment and at least ten such episodes have arrived since the last attempt.
The controller makes at most three training attempts per task, and it does not train when the incumbent's task-success reference has fewer than five trials or the deployment budget is spent.
Two further caps, on total training runs and total episodes per task, bound the overall experiment.

Once the guards pass, the model sees the task memory, trial statistics, previous attempts, the curated instructions with their segment counts, and the code and judge output from up to three recent failures.
It returns \textsc{train} or \textsc{wait} with at most four target instructions.
If the verdict cannot be parsed, the controller waits; an instruction list that cannot be parsed is treated as empty.
Appendix~\ref{app:decision_prompts} gives the prompt.

\begin{table}[!htbp]
\centering
\caption{\textbf{First training request and later decisions by task.} Trial, targets, and data (curated episodes / kept segments) refer to the first training request; arrows show later decisions. Six tasks end with an updated policy. A dash means the controller never requested training.}
\label{tab:first_training_turn}
\footnotesize
\begin{tabular}{@{}p{0.20\linewidth}c p{0.35\linewidth}cc@{}}
\toprule
Task & Trial & Selected target instructions & Data (ep./seg.) & Decision \\
\midrule
Bananas in bin & -- & -- & -- & Base retained \\
Bananas out of bin & 45 & Pick up the banana; place the banana on the table & 43/185 & Adopt \\
Fill the empty bin & -- & -- & -- & Base retained \\
Sort blocks by initial stack position & 60 & Place the green block in the red bowl; place the blue block on the table; pick up the red block & 59/280 & Adopt \\
Return the misplaced object & 60 & Pick up the blue can; place the blue can in the grey bin & 58/220 & Revert$\to$Adopt \\
Recycle cup clutter & 45 & Pick up the coffee cup; place the coffee cup in the black bin & 43/194 & Adopt$\to$Adopt \\
Recycle sort & 45 & Pick up the blue can; place the blue can in the green bin & 39/176 & Revert$\to$Adopt \\
Restack onto the unstacked can & 45 & Pick up the corn can; place the corn can on the tuna can & 42/121 & Revert$\to$Revert \\
Biggest fruit & 45 & Pick up the orange; place the orange into the white bowl & 42/135 & Revert$\to$Adopt \\
Fruits to bowl, except orange & -- & -- & -- & Base retained \\
\bottomrule
\end{tabular}
\end{table}

\section{Adoption arithmetic}
\label{app:adopt}
The evaluation window is 25 episodes, or fewer if the deployment budget runs out.
Execution, recording, and reflection continue as usual during evaluation.
The orchestrator is asked to try the candidate at least once on a relevant skill before falling back, but it is not required to use every target instruction.
A round in which fewer than five episodes contain a candidate call that commits an action is \textsc{starved}.
Both margins in Eq.~\ref{eq:admit} are $0.1$.

The task reference $b$ is computed before training from up to thirty recent episodes served by the incumbent.
For an adopted checkpoint these include its own evaluation window; for the base checkpoint they include only episodes it served outside candidate evaluation windows.
The reference therefore measures the whole system running that checkpoint.

For the prior-skill check, candidate calls are grouped by the instruction strings registered at each earlier adoption.
A group is checked only if it has at least five candidate calls and an incumbent finish rate from before the window; smaller groups are reported but cannot block adoption.
The numerical check matches instructions by exact string, and the semantic reviewer handles paraphrases.
A call counts as finished when its completion check fires or the arm stops moving.
The stillness exit is enabled only for checkpoints trained with held frames.
The semantic reviewer is called only when the exposure floor and both safeguards pass, and a response it cannot parse counts as a veto.

\paragraph{Early stopping.}
After five episodes with candidate calls, a mid-window reviewer may stop evaluation if the candidate has repeatedly failed at its primary skill, the failures are clearly its own, and further trials seem unlikely to change the outcome.
Low task success, an unrelated later failure, or missing evidence is not enough to stop, and unclear or unparseable responses let evaluation continue.
A stopped round never ends in adoption, even if the partial tallies would pass.
If the per-task episode cap is reached mid-window, the round is settled on the episodes collected so far, provided the exposure floor is met.

\section{Fine-tuning}
\label{app:train}
The candidate is optimized on the curated dataset with the policy's flow-matching objective:
\begin{equation}
\mathcal{L}(\theta) =
\mathbb{E}_{(X,\ell)\sim D}\,\mathbb{E}_{t}
\left[\mathcal{L}_{\mathrm{FM}}\!\left(\pi_\theta(o_t,s_t,\ell),a_{t:t+H-1}\right)\right].
\label{eq:obj}
\end{equation}
Here $H=15$ is the prediction horizon.
The dataset stores absolute joint targets and gripper commands.
Before training, joint targets are converted to offsets from the observed joint state, while gripper commands stay absolute; held frames therefore have zero joint offset.
We train LoRA adapters in the language backbone (rank 16) and the action expert (rank 32) and freeze their remaining parameters, while the vision encoder, action input/output projections, and time-conditioning MLP are fully fine-tuned.
Training uses batch size 32, AdamW with gradient clipping at 1.0, and a cosine schedule from $5\times10^{-5}$ to $5\times10^{-6}$.
The number of steps scales with the size of the curated set and is clamped to $[500, 10000]$.
Normalization statistics are recomputed for each curated dataset, and training does not start if they are out of date.
Every run initializes from the pretrained checkpoint.

\section{Agent prompts and decision contracts}
\label{app:decision_prompts}
Quoted text below is verbatim from our prompts; everything else summarizes them.
Curation, training control, and adoption review are separate model calls with different inputs and outputs.
The training controller and adoption reviewer use Gemini 3.8 Flash at temperature $0.2$, with reasoning enabled and a 1024-token output limit.

\paragraph{Skill grouping and visual filtering.}
The text judge receives the task goal, the generated code annotated with executed primitive indices, console output, and the turn's outcome assessment.
Its prompt asks:
\begin{quote}\small
``Your job is to GROUP the marked primitives above into however many distinct LOW-LEVEL SKILLS actually happened in this turn''
\end{quote}
Each returned skill contains \texttt{segment\_indices}, \texttt{instruction}, \texttt{verdict}, and \texttt{reason}, with verdict \texttt{OK}, \texttt{FAILED}, or \texttt{UNCLEAR}.
Every marked primitive must belong to exactly one group, and each label must describe the motion actually executed; for example, closing the gripper on an object is labeled differently from lifting it.
The visual judge receives the instruction and six sampled frames and is told:
\begin{quote}\small
``If the decisive moment for this instruction falls between frames, that is UNCLEAR, not SUCCESS.''
\end{quote}
It returns \texttt{verdict} and \texttt{reason}, with verdict \texttt{SUCCESS}, \texttt{USEFUL\_PARTIAL}, \texttt{NO\_PROGRESS}, \texttt{FAILED}, or \texttt{UNCLEAR}.
For primitive groups, the judge also sees frames from the rest of the turn to check whether an apparent outcome persists; these frames are not added to the training segment.
Missing frames and unparseable responses yield \texttt{UNCLEAR}.

\paragraph{Training request.}
The controller receives curated episode counts (total and new), the most frequent curated instructions with their segment counts, trial statistics, previous attempts, recent failure code and assessments, and task memory.
The prompt opens:
\begin{quote}\small
``You decide whether to spend GPU hours fine-tuning a robot policy for ONE task. Fine-tuning helps when the task's skills are demonstrable from the curated data and the current approach is failing or slow; it wastes money when orchestration already succeeds cheaply or the data is thin/stale.''
\end{quote}
The response follows this format:
\begin{quote}\small\ttfamily
VERDICT: TRAIN or WAIT\par
EXPECT: intended skill improvement and failure to address;\par
\hspace*{1em}on WAIT, evidence that would change the decision.\par
EXPECT\_INSTRUCTIONS: representative primary-capability strings;\par
\hspace*{1em}on WAIT, an empty list.\par
One short paragraph of rationale.
\end{quote}
The prompt asks for one representative string per primary skill, copied from the supplied vocabulary, and the parser keeps at most four.
The semantic reviewer judges these targets by meaning, and supporting motions are listed only if they are themselves the reason to train.
If the instruction list is missing, verification falls back to task-level counts.
Each request is logged with its rationale, targets, and inputs.

\paragraph{Adoption review.}
The reviewer receives the original expectation, the target instructions, per-call records, executed code, and matched outcome assessments.
The prompt states:
\begin{quote}\small
``Mechanical completion (including park-exit) is NOT verified skill success.''\par
``if attribution to the challenger is ambiguous, classify UNCERTAIN.''\par
``Do not count UNCERTAIN as success.''
\end{quote}
The reviewer groups paraphrases and constituent motions into primary skills, separates a successful early skill from a later episode failure, and does not credit the candidate for work completed by scripted recovery.
Explicit gripper operations count as orchestration, so for a placement the reviewer checks whether the candidate reached a usable release pose.
Its response begins with \texttt{ADJUDICATION: CONFIRM} or \texttt{ADJUDICATION: VETO}, followed by its reasoning.

\section{Additional simulation results}
\label{app:simulation_results}
Table~\ref{tab:arms} summarizes the simulation comparison arms, and Table~\ref{tab:sim_full} lists per-task success over all 100 deployment trials and the 50 held-out trials.

\begin{table}[!htbp]
\caption{Simulation comparison arms. Frozen, Fixed, and \abbv{} have matched 100-trial deployment sequences; all methods are tested on held-out initializations.}
\label{tab:arms}
\begin{center}\small
\begin{tabular}{llllc}
\toprule
Arm & Closest setting & Actuation & Adaptation & Deployment \\
\midrule
$\pi_{0.5}$ only & base VLA & policy & none & eval only \\
Frozen-policy baseline & memory-adaptive harness & both & memory & 100 trials \\
Fixed schedule & scheduled policy update & both & memory+policy & 100 trials \\
ASPIRE & agentic skill discovery & programs & skill library & 100 search budget \\
\abbv{} & full method & both & memory+policy & 100 trials \\
\bottomrule
\end{tabular}
\end{center}
\end{table}

\begin{table}[!htbp]
\centering
\caption{\textbf{Simulation success rates (\%).} Deployment uses 100 trials per task and held-out evaluation uses 50. ASPIRE is evaluated on held-out trials only. Bold marks the best result in each group.}
\label{tab:sim_full}
\footnotesize
\setlength{\tabcolsep}{4pt}
\begin{tabular}{@{}p{0.30\linewidth}rrrrrrrr@{}}
\toprule
& \multicolumn{3}{c}{Deployment} & \multicolumn{5}{c}{Held-out} \\
\cmidrule(lr){2-4}\cmidrule(l){5-9}
Task & Frozen & Fixed & Ours & $\pi_{0.5}$ & Frozen & Fixed & ASPIRE & Ours \\
\midrule
Bananas in bin       & 80 & 88 & \textbf{91} & \textbf{100} & 88 & 98 & 88 & 92 \\
Bananas out of bin   & 69 & \textbf{74} & 66 & 74 & \textbf{90} & 88 & 68 & 72 \\
Fill the empty bin       & \textbf{96} & 84 & 94 & 8 & 90 & 82 & \textbf{98} & 80 \\
Sort blocks by initial stack position    & 61 & \textbf{62} & 58 & 0 & 6 & 42 & \textbf{72} & 60 \\
Return the misplaced object    & 69 & 71 & \textbf{82} & 0 & \textbf{92} & 50 & 70 & 86 \\
Recycle cup clutter  & 46 & 50 & \textbf{60} & 0 & 34 & 38 & 0 & \textbf{66} \\
Recycle sort         & 51 & \textbf{56} & 54 & 4 & 30 & \textbf{58} & 4 & 50 \\
Restack onto the unstacked can & 45 & \textbf{51} & \textbf{51} & 2 & 32 & 26 & 54 & \textbf{62} \\
Biggest fruit        & \textbf{79} & 73 & \textbf{79} & 72 & \textbf{90} & 82 & 42 & 76 \\
Fruits to bowl, except orange        & 91 & 94 & \textbf{97} & 20 & \textbf{96} & 94 & 44 & 94 \\
\midrule
Ten-task mean        & 68.7 & 70.3 & \textbf{73.2} & 28.0 & 64.8 & 65.8 & 54.0 & \textbf{73.8} \\
\bottomrule
\end{tabular}
\end{table}

\paragraph{Late-deployment windows.}
Table~\ref{tab:late_windows} reports deployment success over the final $k$ trials for several window lengths, each an equally weighted mean over the ten tasks. \abbv{} has the highest mean in every window.

\begin{table}[!htbp]
\centering
\caption{Deployment success (\%) over the final $k$ trials.}
\label{tab:late_windows}
\small
\begin{tabular}{@{}lccccc@{}}
\toprule
Final $k$ trials & 10 & 20 & 25 & 30 & 50 \\
\midrule
Frozen-policy baseline & 69.0 & 63.5 & 67.6 & 68.7 & 66.2 \\
Fixed schedule        & 68.0 & 65.5 & 67.2 & 68.7 & 68.4 \\
\abbv{}               & \textbf{75.0} & \textbf{75.5} & \textbf{73.6} & \textbf{72.7} & \textbf{73.4} \\
\bottomrule
\end{tabular}
\end{table}

\begin{figure}[!htbp]
\centering
\includegraphics[width=\linewidth]{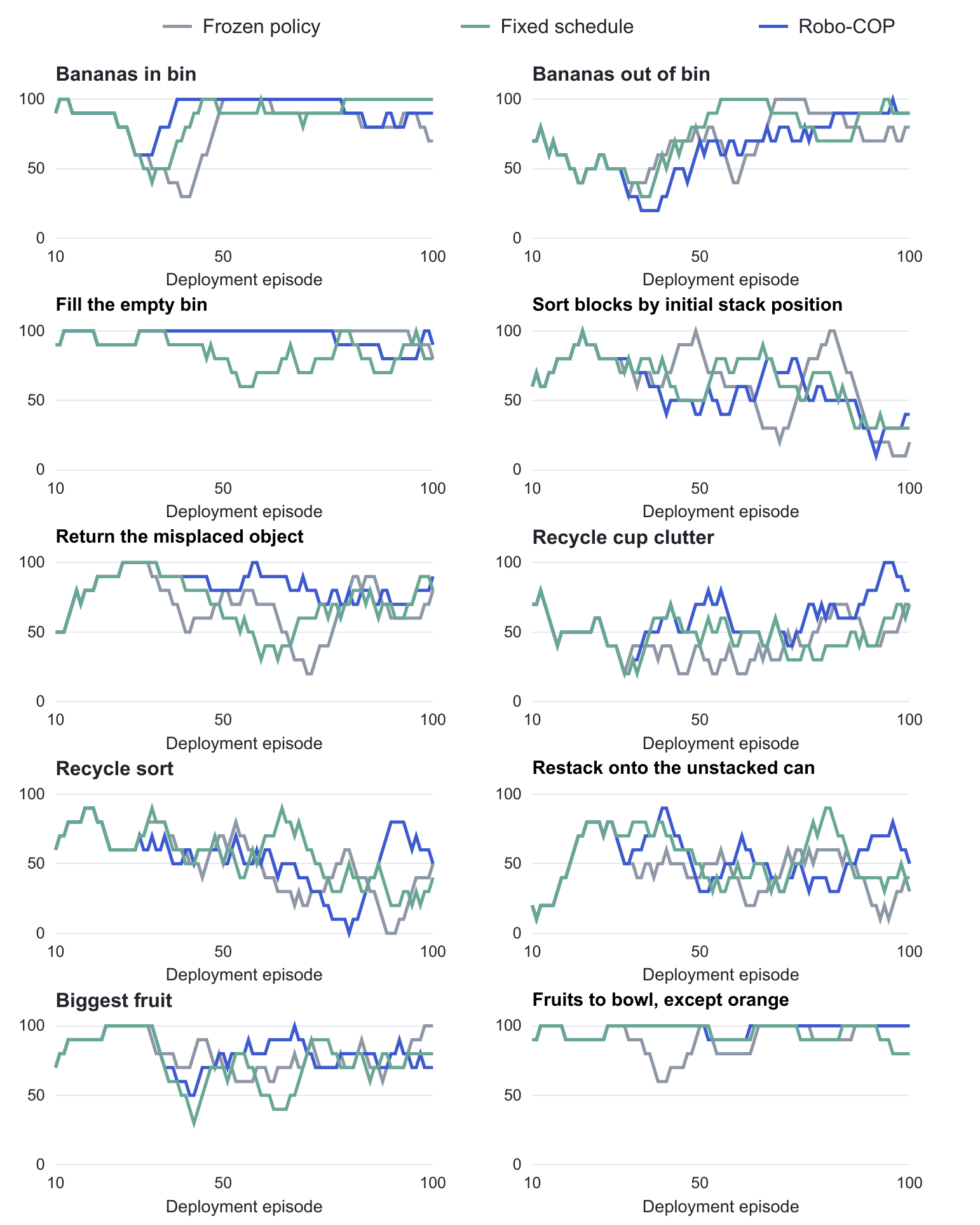}
\caption{\textbf{Trailing 10-episode deployment success on each task.}}
\label{fig:deploy_curves}
\end{figure}

\paragraph{Deployment curves and intervals.}
Figure~\ref{fig:deploy_curves} shows every deployment trajectory in the main-text mean, one run per method and task.
On Sort blocks by initial stack position, all three methods decline late in deployment. Held-out evaluation starts from the memory saved after trial 100, when the frozen-policy baseline was already performing poorly; it falls from 61\% during deployment to 6\% held-out.

\begin{figure}[!htbp]
\centering
\includegraphics[width=0.39\textwidth]{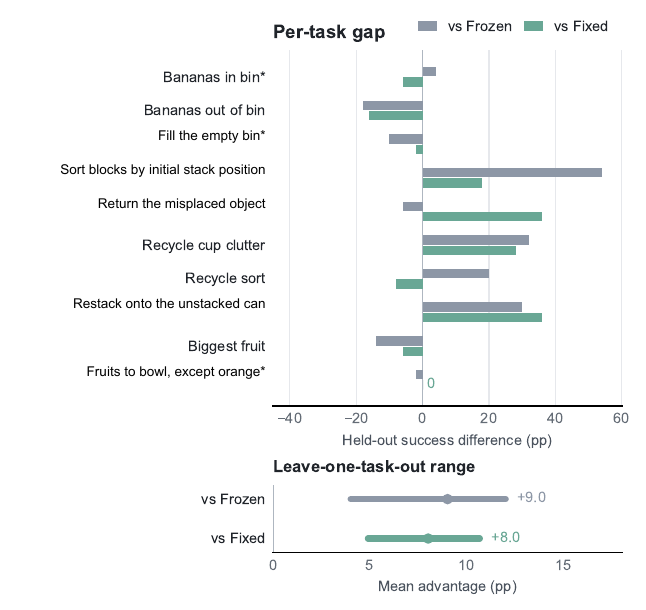}
\caption{\textbf{Held-out task differences.} Bars show \abbv{} minus each baseline, in percentage points. The zero for Fruits to bowl, except orange versus fixed schedule is a tie. The lower panel shows the range of the mean gap when one task is omitted; dots mark the full ten-task means.}
\label{fig:task_robustness}
\end{figure}

\paragraph{Task-level variation.}
Figure~\ref{fig:task_robustness} shows the held-out difference from each baseline on every task, and the range of the mean difference when one task is omitted.
Starred tasks are those where \abbv{} never requested training.

\subsection{Successful episodes and randomized resets}
\label{app:task_visuals}
\begin{figure}[H]
\centering
\includegraphics[width=0.62\linewidth]{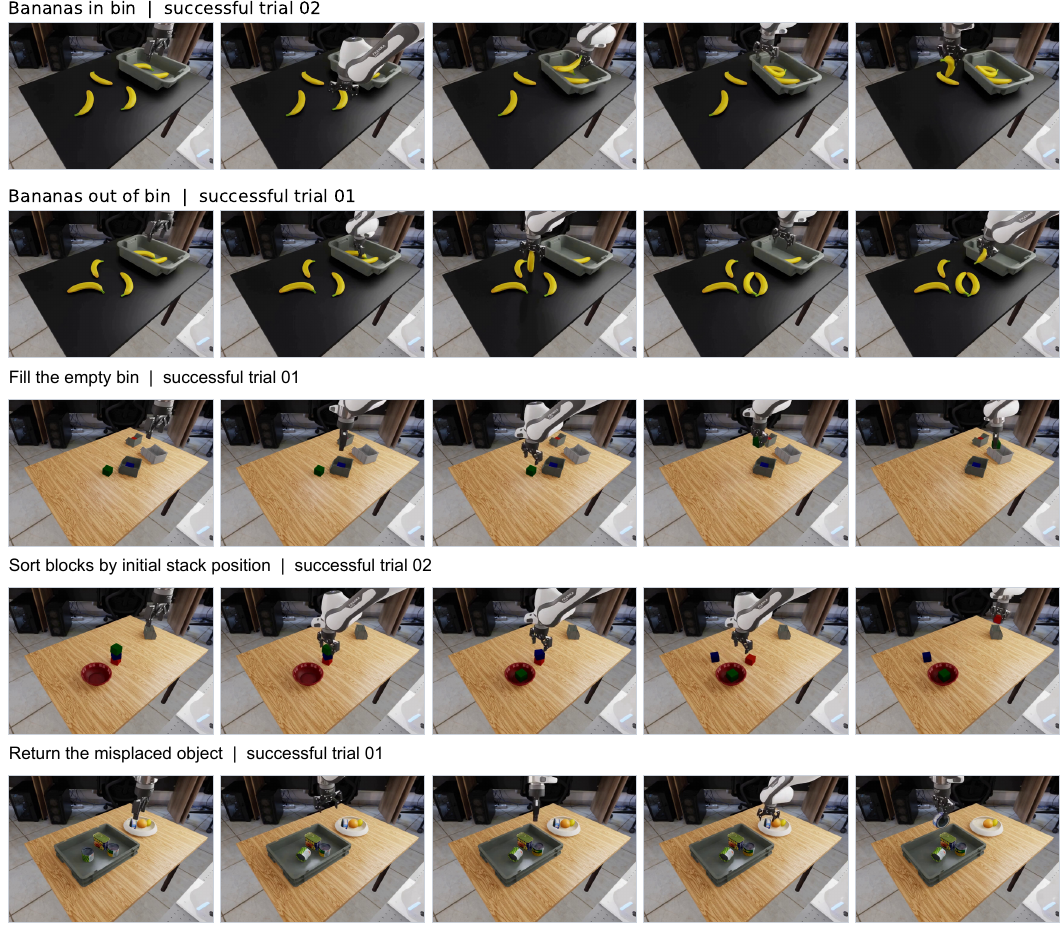}
\caption{\textbf{Successful episodes for tasks 1--5.} Each row shows five frames from one trial, at roughly 2, 25, 50, 75, and 98\% of its duration.}
\label{fig:success_1to5}
\end{figure}

\begin{figure}[H]
\centering
\includegraphics[width=0.62\linewidth]{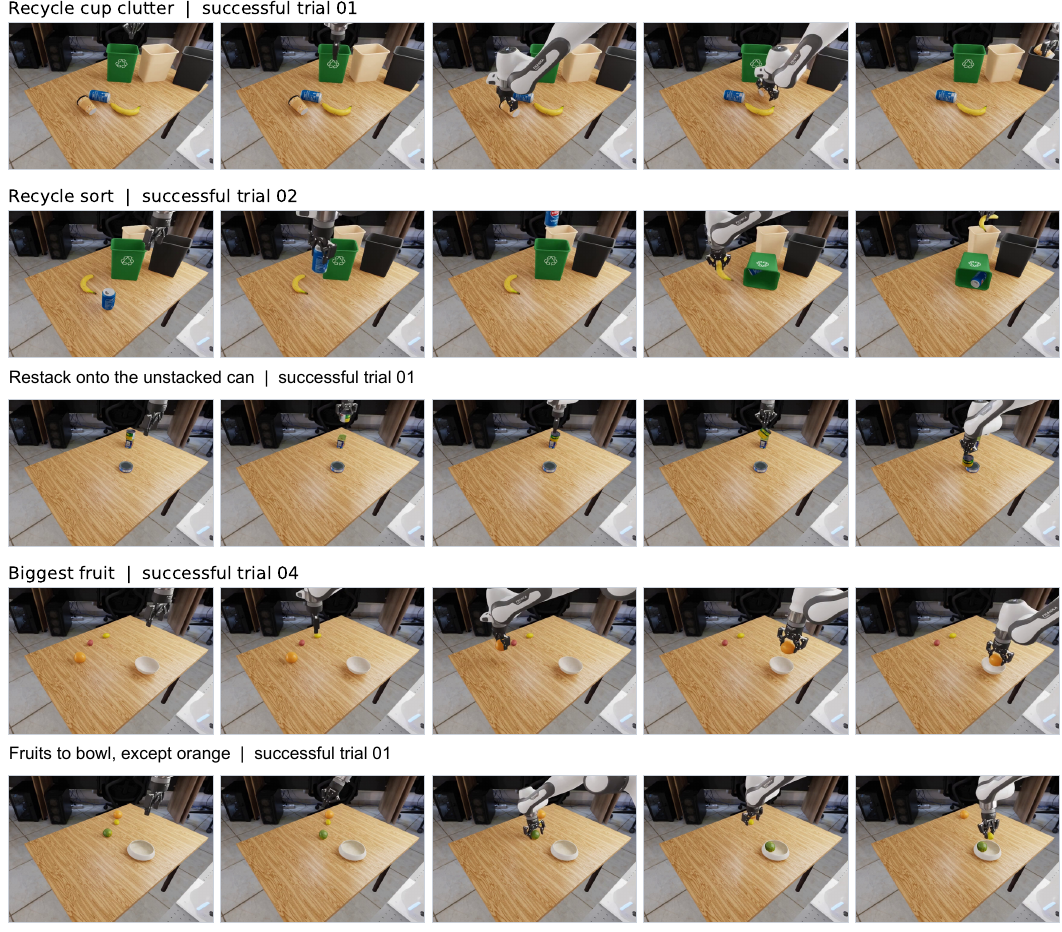}
\caption{\textbf{Successful episodes for tasks 6--10.} Frames are sampled as in Figure~\ref{fig:success_1to5}.}
\label{fig:success_6to10}
\end{figure}

\begin{figure}[H]
\centering
\includegraphics[width=0.64\linewidth]{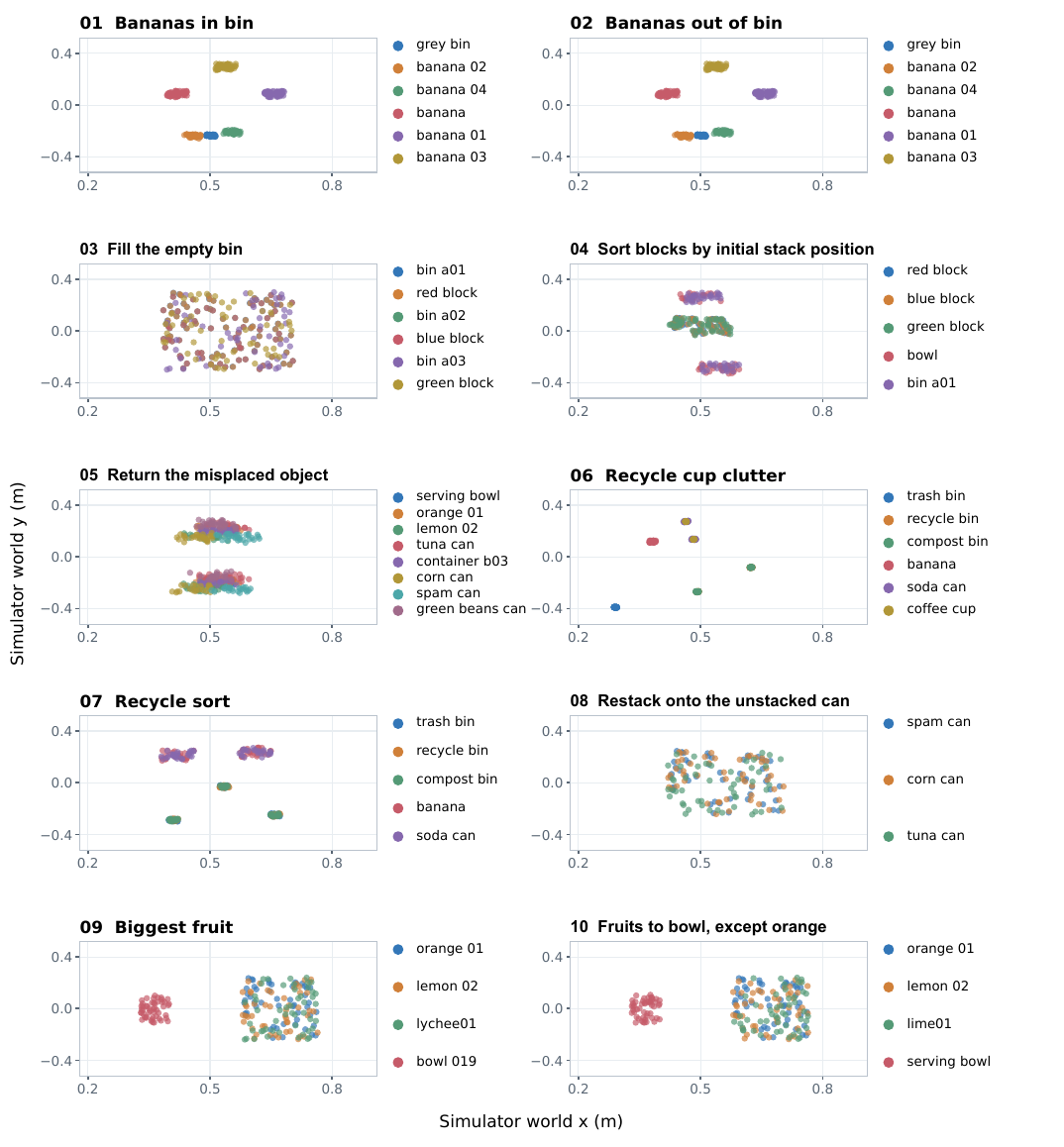}
\caption{\textbf{Initial object positions after 50 seeded resets per task, in simulator world coordinates (m).} Colors distinguish objects within each task.}
\label{fig:resets}
\end{figure}

Figures~\ref{fig:success_1to5} and~\ref{fig:success_6to10} show one successful deployment episode per task, five frames each.
Figure~\ref{fig:resets} plots the initial object positions over 50 seeded resets per task.

\section{Real-world executions}
\label{app:real_world}
\begin{figure}[H]
\centering
\includegraphics[width=0.80\linewidth]{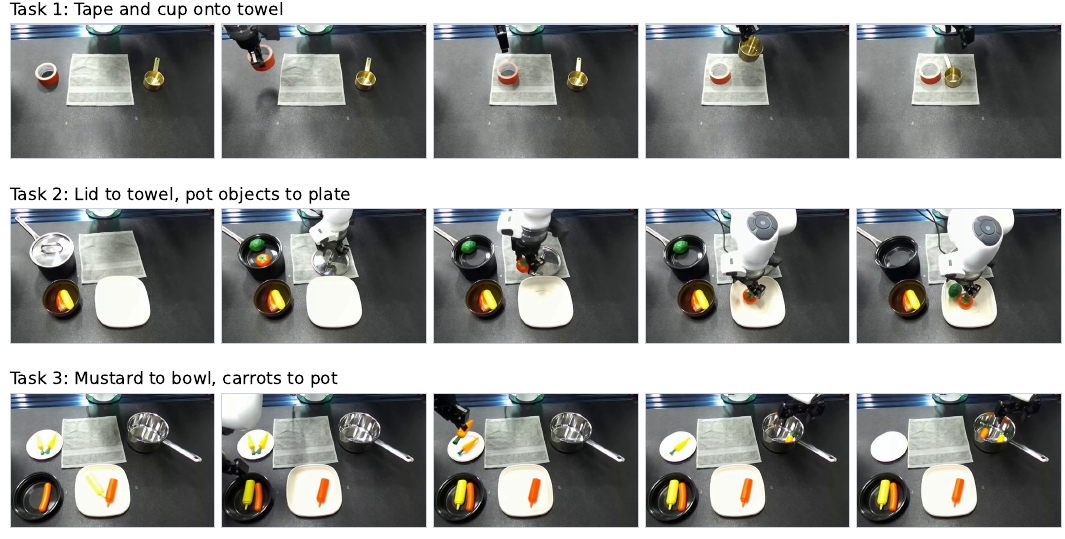}
\caption{\textbf{Successful real-world executions of the three tasks.} Frames are shown in temporal order.}
\label{fig:real_world_sequences}
\end{figure}

Figure~\ref{fig:real_world_sequences} shows successful executions of the three real-world tasks in Section~\ref{sec:real}.

\end{document}